\documentclass[letterpaper, 10 pt, conference]{ieeeconf}  

\usepackage{times}
\usepackage{epsfig}
\usepackage{graphicx}
\usepackage{amsmath}
\usepackage{amssymb}
\usepackage{multirow}
\usepackage{tabularx, booktabs}
\usepackage{xcolor}

\usepackage[pagebackref=true,breaklinks=true,letterpaper=true,colorlinks,bookmarks=false]{hyperref}

\IEEEoverridecommandlockouts                             

\title{\LARGE \bf
Learning 2D to 3D Lifting for Object Detection in 3D \\
for Autonomous Vehicles
}

\author{Siddharth Srivastava$^{1}$, Frederic Jurie$^{2}$ and Gaurav Sharma$^{3}$
\thanks{$^{1}$Siddharth Srivastava is with Indian Institute of Technology Delhi, India
        {\tt\small eez127506@ee.iitd.ac.in}}%
\thanks{$^{2}$Frederic Jurie is with Normandie Univ., UNICAEN, ENSICAEN, CNRS, France
        {\tt\small frederic.jurie@unicaen.fr}}%
\thanks{$^{3}$Gaurav Sharma is with NEC Labs America, USA
         {\tt\small grvsharma@gmail.com}}
}

\begin{document}

\maketitle

\def\etal{et al.\hspace*{1mm}}
\def\etc{etc.\hspace*{1mm}}
\def\ie{i.e.\hspace*{1mm}}
\def\eg{e.g.\hspace*{1mm}}
\def\cf{cf.\hspace*{1mm}}
\def\vs{vs.\hspace*{1mm}}
\def\L{\mathcal{L}}
\def\T{\mathcal{T}}
\def\x{\textbf{x}}
\def\R{\mathbb{R}}

\begin{abstract}
We address the problem of 3D object detection from 2D monocular images in autonomous driving scenarios. We propose to lift the 2D images to 3D representations using learned neural networks and leverage existing networks working directly on 3D data to perform 3D object detection and localization. We show that, with carefully designed training mechanism and automatically selected minimally noisy data, such a method is not only feasible, but gives higher results than many methods working on actual 3D inputs acquired from physical sensors. On the challenging KITTI benchmark, we show that our 2D to 3D lifted method outperforms many recent competitive 3D networks while significantly outperforming previous state-of-the-art for 3D detection from monocular images. We also show that a late fusion of the output of the network trained on generated 3D images, with that trained on real 3D images, improves performance. We find the results very interesting and argue that such a method could serve as a highly reliable backup in case of malfunction of expensive 3D sensors, if not potentially making them redundant, at least in the case of low human injury risk
autonomous navigation scenarios like warehouse automation.
\end{abstract}

\section{Introduction}

We address the important problem of object detection in 3D data while only using monocular images at inference. Traditionally, two approaches have been widespread for 3D object detection problems. First is to detect objects in 2D using monocular images and then infer in 3D \cite{chen2016monocular, xu2018multi, li2018deep}, and second is to use 3D data (\eg LiDAR) to detect bounding boxes directly in 3D \cite{chen2017multi}. However, on 3D object detection and localization benchmarks, the methods based on monocular images significantly lag behind the latter, limiting their deployment in practical scenarios. A reason for such a disparity in performance is that methods based on monocular images attempt at implicitly inferring 3D information from the input. Additionally, availability of depth information (derived or explicit), greatly increases the performance of such methods \cite{chen2017multi}. Moreover, for 3D networks to work at test time, a limitation is the need to deploy expensive (thousands of dollars) and bulky (close to a kilogram) 3D scanners \cf cheaper and lighter 2D cameras. Hence, a monocular image based 3D object detection method closing the gap in performance with the methods requiring explicit 3D data will be highly practical. 

\begin{figure}
\centering
\includegraphics[width=\columnwidth,trim=0 175 360 0,clip]
{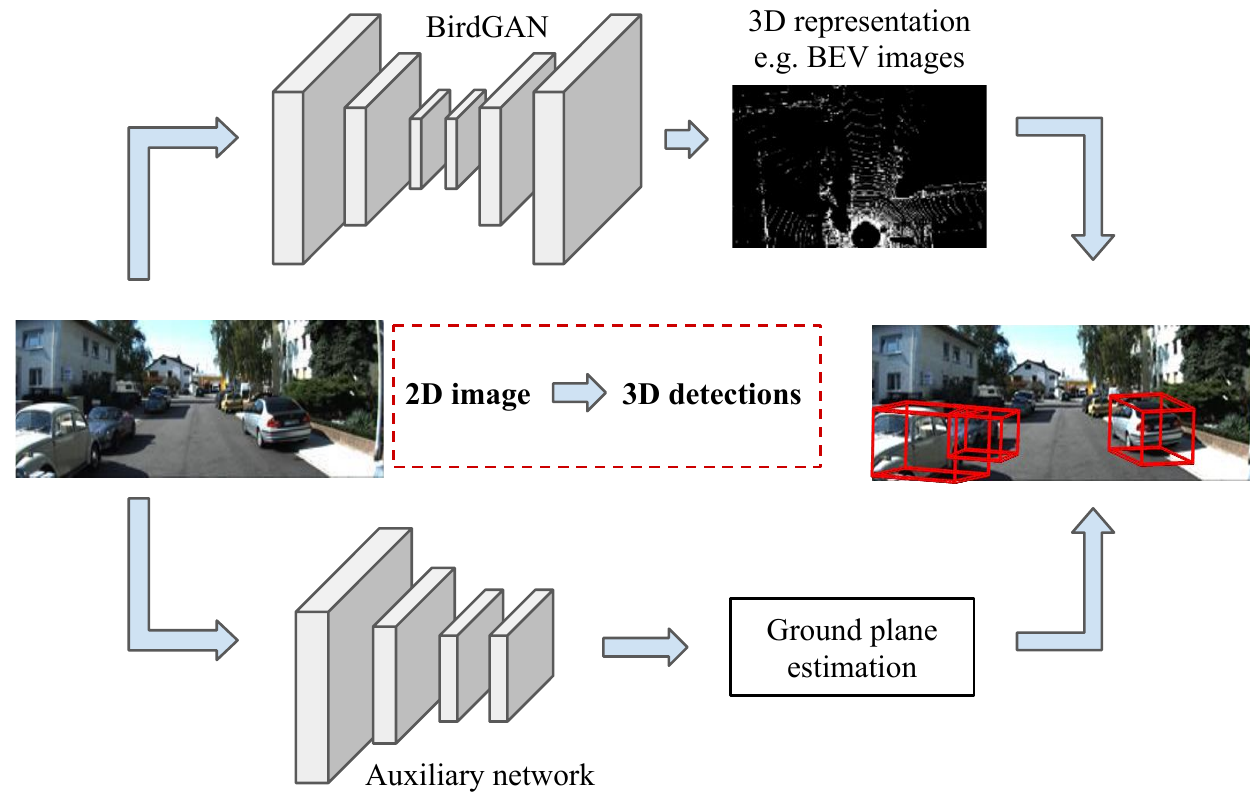}

\caption{We aim to do 3D detection from 2D monocular images, by generating (i) 3D representations using state-of-the-art GANs and (ii) 3D data for ground plane estimation using recent 3D networks. We show that it is possible to achieve competitive 3D detection without having actual 3D data at test time.}
\end{figure}

In this paper, we show that we can learn a mapping, leveraging existing 3D data, and use it to perform object detection in 3D at test time without the need for explicit 3D data. We show that it is possible to utilize 3D data, collected once, to train a network to \emph{generate} 3D information at test time from 2D images, which can be used as a drop-in replacement to many object detection pipelines, and still provide surprisingly good results. In particular, we target the Bird's Eye View (BEV) and 3D Object Detection challenges of KITTI's evaluation benchmark and show that with 3D information generated at test time from 2D images, we can achieve better or comparable results to numerous recent and competitive techniques working directly on 3D data. We believe the results are of importance as (i) the efforts that are directed towards collecting high quality 3D data can help in scenarios where explicit 3D data cannot be acquired at test time. (ii) the method can be used as a plug-and-play module with any existing 3D method which works with BEV images, allowing operations with seamless switching between RGB and 3D scanners while leveraging the same underlying object detection platform. 

The presented work makes use of progress in interpretation of 2D scans in 3D, such as 3D reconstruction from single images \cite{fan2017point, yan2016perspective} and depth estimation \cite{godard2017unsupervised}. To the best of our knowledge, we are the first to generate (two different variants of) BEV image from a single RGB image, using state-of-the-art image to image translation models, and use it for object detection using existing 3D CNNs. The results show that we significantly outperform the state-of-the-art monocular image based 3D object detections while also performing better than many recent methods requiring explicit 3D data. Finally, while the performance of a method with generated data is expected to be at most as good as the underlying network, we show that by fusing the outputs of the base network and the network trained on generated data, the performance of the base network is further improved,  outperforming the methods based on 3D data captured using scanners. 
\section{Related Work}
\noindent
\textbf{Object Detection in 3D:} 
Object detection in 3D is one of the main tasks of 3D scene understanding. Many works have addressed 3D object detection using 3D data like LiDAR images \cite{chen2017multi, beltran2018birdnet,xiang2015data} and stereo images \cite{chen20183d}, while some have also used only monocular images \cite{chen2016monocular,song2015joint}.  The approaches for 3D object detection vary from proposing new neural network architectures, \eg BirdNet
\cite{beltran2018birdnet}, MV3D \cite{chen2017multi}, to novel object representations such as the work on 3DVP \cite{xiang2015data}. Some works also utilize other modalities along with 3D, such as corresponding 2D images \cite{chen2017multi} and structure from motion \cite{7780838}. Among the neural network based approaches, many competitive approaches follow the success of 2D object detection methods and are based on 3D proposal networks and classifying them, \eg MV3D (multi-view 3D) \cite{chen2017multi},  AVOD\cite{ku2018joint}. 

We build on such recent architectures which work directly with 3D representations, i.e., previous
works that took multiview projections of the 3D data to use with 2D image networks followed by
fusion mechanisms \cite{su2015multi}. However, instead of feeding them with real 
3D data we use generated data as input. Since the two architectures we use as our base networks, \ie
BirdNet \cite{beltran2018birdnet} and MV3D \cite{chen2017multi}, take inputs of different
nature we propose appropriate generation networks and a carefully designed training data
processing pipeline.
\vspace{0.5em}\\
\textbf{Inferring 3D using RGB images:} 
Among methods inferring 3D information from RGB images, \cite{wu2016single} work on predicting 2D
keypoint heat maps and 3D objects structure recovery.  \cite{saxena2009make3d} use single RGB image
to obtain detailed 3D structure using MRFs on small homogeneous patches to predict plane parameters
encoding 3D locations and orientations of the patches.  \cite{pavlakos2017coarse} learn to predict 3D
human pose from single image using a fine discretization of the 3D space around the subject and
predicting per voxel likelihoods for each joint, and using a coarse-to-fine scheme.

\begin{figure*}[ht]
\centering
	\includegraphics[width=.95\textwidth,trim=0 220 30 0, clip]{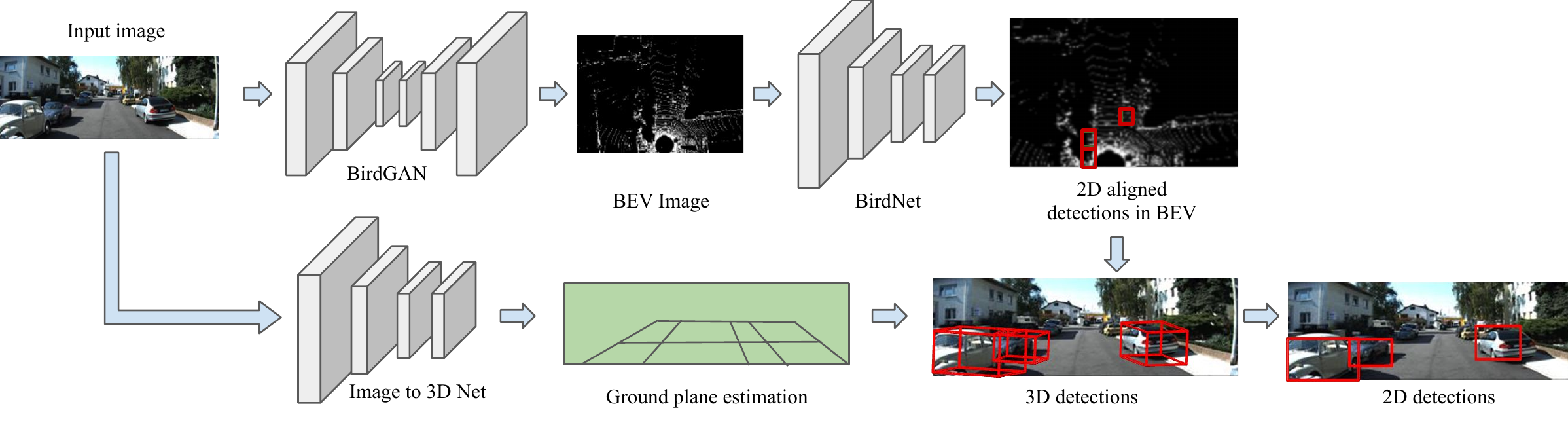}
	\caption{Proposed pipeline with BirdNet. A GAN based generator translates the 2D RGB image into BEV image compatible with the BirdNet architecture. An RGB to 3D network, like the Perspective Transformer Network \cite{yan2016perspective}, gives us 3D information for ground planes estimation. The BEV detections are then converted to 3D detections using the ground plane estimation.}
	\label{figBirdNet}
\end{figure*}

\noindent
\textbf{Generating 3D data from 2D:} Many works have proposed variants of generative models for 3D data generation.
\cite{wu2016learning} use Generative Adversarial Networks (GANs) to generate 3D objects using volumetric networks, extending the
vanilla GAN and VAE GAN to 3D. \cite{gadelha20173d} propose projective generative adversarial networks (PrGAN) for obtaining 3D structures from multiple 2D views. 
\cite{park2017transformation} synthesize novel views from a single image by inferring
geometrical information followed by image completion, using a combination of adversarial and
perceptual loss. \cite{yan2016perspective} propose Perspective Transformer Nets (PTNs), an
encoder-decoder network with a novel projection loss using perspective transformation, for learning
to use 2D observations without explicit 3D supervision. \cite{zhu2018learning} generate 3D models
with an enhancer neural network extracting information from other corresponding domains (e.g.
image). \cite{yang20173d} uses a GAN to generate 3D objects from a single depth image,
by combining autoencoders and conditional GAN.
\cite{smith2017improved} uses a GAN to generate 3D from 2D images, and perform shape completion from
occluded 2.5D views, building on \cite{wu2016learning} using Wasserstein objective.
\vspace{0.5em}\\
\textbf{Image to image translation:} Our work addresses the specific task of 3D object detection by translating RGB images to BEV. Image translation has recently received attention for style transfer applications, \eg pix2pix \cite{isola2017image} or the recent work of \cite{zhu17rowardmultimodal}.  While perhaps being less challenging
than a full and accurate 3D scene generation, 3D object detection is still a very challenging and relevant task for
autonomous driving use cases. Here, we generate 3D data as an intermediate step, but instead of focusing
on the quality of the generated 3D data as in \cite{isola2017image,zhu17rowardmultimodal}, we design and evaluate our method directly on the task of
3D object detection from monocular images.
\section{Approach}
The proposed approach aims at generating 3D data from 2D images for performing 3D object detection in the generated data with 3D object detectors. We generate BEV images directly from a single RGB image, (i) by designing a high fidelity GAN architecture, and (ii) by carefully curating a training mechanism, which includes selecting of minimally noisy data for training. 

Since training GAN based networks is hard, we initially explored the idea of obtaining 3D point clouds image to depth networks \cite{godard2017unsupervised}, and project BEV from them. However, the generated BEVs were sparse and did not have enough density/information in the object regions. This motivated the current approach of directly generating BEV using GAN. 

The proposed method can work with different variants of the BEV image considered. Specifically, we show results with two recent competitive 3D object detection networks, BirdNet~\cite{beltran2018birdnet} and MV3D~\cite{chen2017multi}, which originally take different formats of BEV inputs obtained form 3D LiDAR inputs. Both of the networks process the LiDAR input to obtain (i) Bird's Eye View (BEV) images, of two different variations, and (ii) 3D point clouds for ground plane estimation. Additionally, MV3D also takes the corresponding front view and RGB images as input.  The BEV images generated from 2D images by the proposed method are effective with such 3D detection algorithm, both at train and test times. While, BirdNet is based on processing BEV input, MV3D is takes multi-modal input. We show results on both, and hence demonstrate that the proposed method is general and is capable of working across a spectrum of methods using (variations of) BEV input. 
We now gives details of our generation network, and our training data processing, followed by the details of two instantiations of our method.

\begin{figure*}
\centering
	\includegraphics[width=.95\textwidth,trim=0 220 0 0,clip]{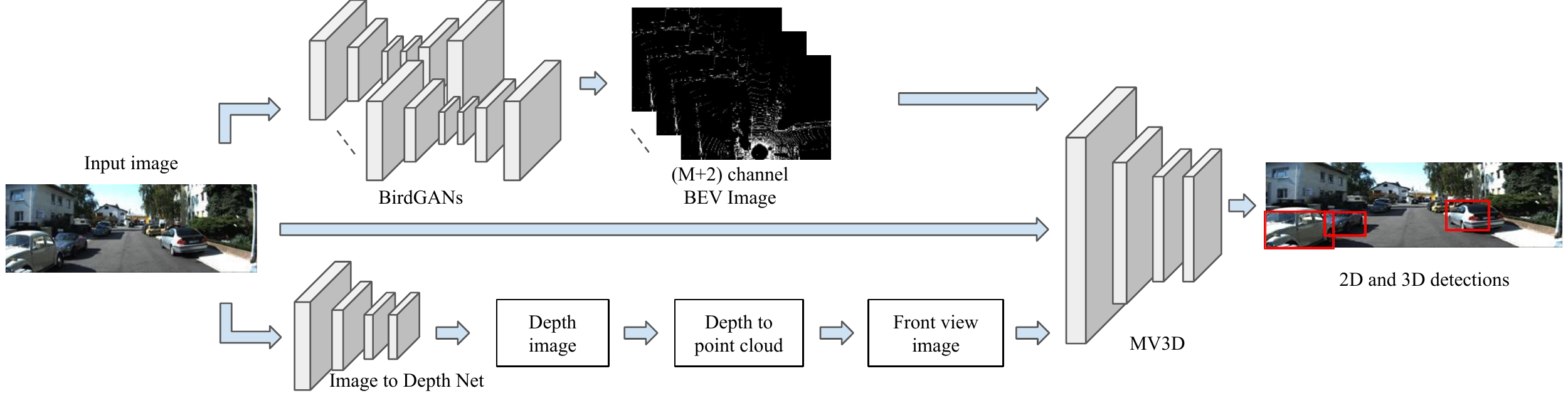}
	\caption{Proposed pipeline with MV3D. In the case of MV3D, multiple GAN based generators independently translate the 2D RGB image into the different channels of the MV3D compatible BEV image. In addition auxiliary network is used for Front View (FV) image generation from RGB image. All three, \ie RGB, FV and BEV images, are then fed to the MV3D architecture to predict in 3D.}
	\label{figMV3D}
\end{figure*}

\subsection{Generating BEV images from 2D images (BirdGAN) \label{sec:birdgan}}
The network for generating the BEV images from input 2D RGB images are based on the Generative Adversarial Networks for image to image translation \cite{isola2017image}. GANs have become very popular recently and here we use them for turning 2D RGB images to BEV images containing 3D information about the scene. Specifically, the image translation BirdGAN network consists of a VGG-16 CNN to encode the images, followed by DCGAN \cite{radford2015unsupervised} conditioned over the encoded vector to generate the BEV image. The full BirdGAN pipeline is trained end-to-end using the paired monocular and BEV images available.

\begin{figure}
    \centering
	\includegraphics[width=.9\columnwidth,trim=0 130 270 0,clip]{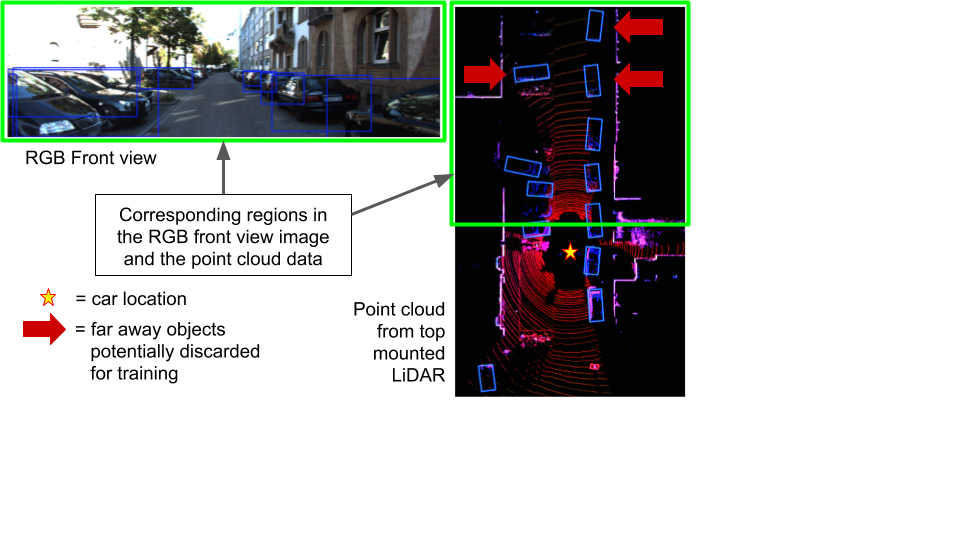}
	\caption{The RGB image only shows the front view while the top mounted LiDAR point cloud also have data from the back and sides of the car. We crop the LiDAR point cloud appropriately so that only the corresponding information in the two modalities remain. We also prune out far away BEV points, as they are highly occluded in the RGB image, potentially loosing some objects \eg those highlighted with red arrows.}
	\label{figRGBLiDAR}
\end{figure}

The quality of data used to train the GAN makes a big difference in the final performance. With this in mind, we propose and experiment with two methods for training the GAN for generating BEV images. In the first, we take all the objects in the scene, \ie the whole image, to generate the BEV image, while in the second we take only the `well-defined' objects in the scene, i.e., those closest to the camera. The former is the natural choice which makes use of all the data available, while the latter is motivated by the fact that the point clouds become relatively noisy, and possibly uninformative for object detection, as the distance increases due to very small objects and occlusions. In the second case, while a significant amount of data is discarded (\eg the objects highlighted with red arrows in the BEV image on the right in Fig.~\ref{figRGBLiDAR}), the quality of retained data is better as the nearby objects are bigger and have fewer occlusions etc., especially in the RGB image. In both of the cases, we work with the RGB images and corresponding LiDAR clouds of the area around the car (Fig.~\ref{figRGBLiDAR}).

\subsection{BirdNet 3D object detection}
\vspace{-0.6em}
The default BirdNet \cite{beltran2018birdnet} pipeline uses a $3$ channel Bird's Eye View (BEV) image consisting of height, density and intensity of the points as the main input. In addition to the BEV image input, BirdNet also requires ground plane estimation for determining the height of the 3D bounding boxes. Both of these inputs are normally extracted from the full LIDAR point cloud which is the main input to the system.
\vspace{-0.15em}

In the proposed method ((Fig.~\ref{figBirdNet}), we generate the two inputs, \ie the BEV image and the 3D point cloud using two neural networks learned on auxiliary data, respectively. The first network is the GAN based network explained in the previous section (Sec.~\ref{sec:birdgan}). It takes the RGB image as input and outputs the $3$ channel BEV image.  The $3$ channels of the BEV image in this case are the height, density and intensity of the points. 
\vspace{-0.15em}

The second network reconstructs a 3D model using the RGB image as input. The 3D reconstruction network takes the $3$ channel RGB image as input and generates either the point clouds or their voxelized version as the 3D model. The generated 3D model is then used to obtain the ground estimation for constructing the 3D bounding boxes around the detected objects.

\subsection{MV3D as base architecture} \label{subsec:mv3darch}
MV3D \cite{chen2017multi} takes three inputs, the RGB image, the LiDAR Front View \cite{li2016vehicle} and the BEV image. It differs from BirdNet in the format of BEV input it accepts, while BirdNet takes a $3$ channel BEV image, \ie height, intensity and density, MV3D pre-processes the height channel to encode more detailed height information. It divides the point cloud into $M$ slices and computes a height map for each slice giving a BEV image of $M+2$ channels. Hence, to generate the appropriate input for MV3D, we use multiple independently trained BirdGANs to generate the $M$ height channels of the BEV image. We also experimented with directly generating the $M+2$ channel BEV image, but the results indicated that independently trained GANs provided better results. We use the RGB image to obtain the corresponding depth image using a neural network \cite{godard2017unsupervised}, and use the depth map image to obtain the 3D point cloud, for constructing the LiDAR front view. Following \cite{chen2017multi, li2016vehicle} we project the information on to a cylindrical plane  obtaining the front view. We finally feed all these three inputs, \ie the BEV image, the front view image and the RGB image, to the MV3D network to obtain 3D object detections (Fig.~\ref{figMV3D} illustrates the full pipeline).

\subsection{Ground plane estimation}
\vspace{-0.3em}
In the proposed pipelines, the ground plane estimation is needed in both cases. BirdNet uses the ground plane, \ie the bottom-most points, to estimate the height of the object for constructing the 3D bounding boxes. MV3D obtains the 3D localizations by projecting the 3D bounding boxes to the ground plane. The ground plane estimation is an important step here, especially for MV3D, as it governs the size of the projected objects on the BEV impacting the quality of 3D object localization.  

There are two ways to obtain the  ground plane, (i) by reconstructing a 3D model from a single RGB image using techniques such as Perspective Transformer Network (PTN)\cite{yan2016perspective}, Point Set generation \cite{fan2017point}, depth estimation \cite{godard2017unsupervised} \etc and estimate the ground plane, or (ii) using the image to directly estimate the ground plane without transforming the image to 3D \cite{song2015joint,cherian2009accurate,wang2014color}. The quality of the latter case usually requires explicit presence of strong 2D object proposals or texture/color pattern. \cite{murthy2017shape} also noted that presence of shape prior for ground plane estimation significantly improves the performance on 3D object localization. Recently it was also shown that a class specific end-to-end learning framework, even on synthetic dataset, could provide accurate pose and shape information. Executing such networks at test time is fast, as it usually involves a single forward pass. Therefore, we choose the former paradigm with Perspective Transformer Network and reconstruct the 3D object/scene. The ground plane is then estimated by fitting a plane using RANSAC~\cite{choi2014robust}. 

\section{Experiments}
\noindent
\textbf{Dataset:}
We evaluate our method on the standard challenging KITTI Object Detection Benchmark \cite{geiger2012we}. The dataset consists of $7,481$ training images and $7,518$ images for testing, however, the test set is not publicly available. We split the train set into training and validation used in \cite{chen20153d}, and report results with these splits, as done by many recent works. While we focus on techniques which use Bird's Eye View (BEV) images for detecting objects, we also compare against published state-of-the-art results to demonstrate the competitiveness of the proposed method.

\noindent
\textbf{Training data for BirdGAN:}
We use RGB image as input for training BEV GAN for BirdNet with the $3$ channels, height, intensity and density, BEV image as required by BirdNet. In case of MV3D, the BEV is different than Birdnet; the height channel consists of $M$ channels which are produced by slicing the point cloud in $M$ slices along the height dimensions. We experiment with a single GAN to generate $M$+$2$ channels as output and as well as multiple independent GANs for each of the slices. 

The results are reported by training BirdGAN on two types of training data for both the above cases.
\vspace{0.3em} \\
\textit{(w/o clipping)} --- We use the data in the field of view of RGB images \ie $90^\circ$ in the front view. This setting is referred to as without clipping.
\vspace{0.3em} \\
\textit{(clipping)} --- In KITTI dataset, the objects that are far, are difficult to detect mostly due to occlusion~\cite{song2015joint}. Using such data could hinder training the GAN, as then the GAN is expected to generate objects which are not clearly visible in the RGB image. We experiment with using only the nearby objects for training the GANs, \ie we remove the BEV image data corresponding to points which are more than $25$ meters away and use these modified BEV images to train the GAN based translation model. This data setting is referred to as (with) clipping.
\\
\noindent
\textbf{Parameter Settings:} For training MV3D, BirdNet, PTN and image to depth networks, we use the the default experimental settings from the respective publications.  
\\
\noindent
\textbf{Proposed methods:}
In the experimental results, following notations are used for various proposed methods.
\vspace{0.3em}\\
\emph{Ours (w/o clipping)-BirdNet} --- Generated BEVs are used for training and testing the BirdNet. The GAN is trained on data without clipping. For constructing the training pairs for BirdGAN with BirdNet, the input consists of RGB images while the output consists of BEV images generated using the technique of BirdNet. Once the BirdGAN is trained, the BirdNet is trained using BirdGAN generated BEVs from input RGB images. At test time, the input image is again mapped to a BEV image by BirdGAN and passes to BirdNet for further processing. 
\vspace{0.3em}\\
\emph{Ours (w/o clipping)-MV3D} --- This refers to using BirdGAN trained on data without clipping. Here the output of BirdGAN is of $M$+$2$ where the ground truth BEV is generated using the technique of \cite{chen2017multi}.
\vspace{0.3em}\\
\emph{Ours (clipping)-BirdNet} --- This setting refers to using clipped training for training BirdGAN and its subsequent use to train and test BirdNet described above.
\vspace{0.3em}\\ 
\emph{Ours (clipping)-MV3D} --- Similar to the previous case with the training and testing pairs consist of clipped data.

\begin{table}
	\begin{center}
		\begin{small}
			\addtolength{\tabcolsep}{-0.1pt}
			\resizebox{\linewidth}{!}{
			\begin{tabular}{c c c c c c c c}
				\hline
				\multirow{2}{*}{Method}& \multirow{2}{*}{Data}& \multicolumn{3}{c}{IoU=0.5} & \multicolumn{3}{c}{IoU=0.7} \\
				\cline{3-8}
				& & Easy & Moderate & Hard & Easy & Moderate & Hard \\
				\hline
				Mono3D~\cite{chen2016monocular} & Mono & 30.5 & 22.39 & 19.16 & 5.22 & 5.19 & 4.13 \\
				3DOP~\cite{chen20153d} & Stereo & 55.04 & 41.25 & 34.55 & 12.63 & 9.49 & 7.59\\
				Xu et. al \cite{xu2018multi}                      & Mono           & 55.02                  & 36.73                   & 31.27                   & 22.03         & 13.63         & 11.60         \\ \hline
				\hline
				BirdNet~\cite{beltran2018birdnet} & LIDAR & 90.43 & 71.45 & 71.34 & 72.32 & 54.09 & 54.50 \\
				DoBEM~\cite{yu2017vehicle} & LIDAR & 88.07 & 88.52 & 88.19 & 73.09 & 75.16 & 75.24 \\
				VeloFCN~\cite{li2016vehicle} & LIDAR & 79.68 & 63.82 & 62.80 & 40.14 & 32.08 & 30.47 \\
				MV3D (BEV+FV)~\cite{chen2017multi} & LIDAR & 95.74 & 88.57 & 88.13 & 86.18 & 77.32 & 76.33 \\
				MV3D (...+RGB)~\cite{chen2017multi} & LI+Mo & 96.34 & 89.39 & 88.67 & 86.55 & 78.10 & 76.67 \\
				Frustum PointNets~\cite{qi2017frustum} & LIDAR & - & - & - & 88.16 & 84.02 & 76.44 \\
				\hline \hline
				Ours \\
				(w/o clipping)-BirdNet & Mono & 58.40 & 49.54 & 48.20 & 45.60 & 31.10 & 29.54 \\
				(w/o clipping)-MV3D & Mono & 71.35 & 47.43 & 43.25 & 58.70 & 38.20 & 36.56 \\
				(clipping)-BirdNet & Mono & 84.40 & 64.18 & 58.70 & 68.2 & 42.1 & 36.1 \\
				(clipping)-MV3D & Mono & 90.24 & 79.50 & 80.16 & 81.32 & 68.40 & 60.13 \\
				\hline
			\end{tabular}
			}
		\end{small}
		\caption{{\bf 3D localization performance:} Average Precision (AP$_{\text{loc}}$, \%) of bird's eye view boxes on KITTI \emph{validation} set. For Mono3D and 3DOP, we use the results with 3D box regression from \cite{chen2017multi}}
		\label{tab:ap_loc_val}
		\vspace{-1em}
	\end{center}
\end{table}

\begin{table*}
	\begin{center}
		\begin{small}
			\addtolength{\tabcolsep}{-0pt}
			\resizebox{0.85\linewidth}{!}{
			\begin{tabular}{ c c c c c c c c c c c}
				\hline
				\multirow{2}{*}{Method}& \multirow{2}{*}{Data}& \multicolumn{3}{c}{IoU=0.25} & \multicolumn{3}{c}{IoU=0.5} & \multicolumn{3}{c}{IoU=0.7} \\
				\cline{3-11}
				& & Easy & Moderate & Hard & Easy & Mod & Hard & Easy & Mod& Hard \\
				\hline
				Mono3D~\cite{chen2016monocular} & Mono & 62.94 & 48.2 & 42.68 & 25.19 & 18.2 & 15.52 & 2.53 & 2.31 & 2.31 \\
				3DOP~\cite{chen20153d} & Stereo & 85.49 & 68.82 & 64.09 & 46.04 & 34.63 & 30.09 & 6.55 & 5.07 & 4.1 \\
				\hline
				\hline
				BirdNet~\cite{beltran2018birdnet} & LIDAR & 93.45 & 71.43 & 73.58 & 88.92 & 67.56 & 68.59 & 17.24 & 15.63 & 14.20 \\
				
				VeloFCN~\cite{li2016vehicle} & LIDAR & 89.04 & 81.06 & 75.93 & 67.92 & 57.57 & 52.56 & 15.20 & 13.66 & 15.98 \\
				MV3D (BEV+FV)~\cite{chen2017multi} & LIDAR & 96.03 & 88.85 & 88.39 & 95.19 & 87.65 & 80.11 & 71.19 & 56.60 & 55.30 \\
				MV3D (BEV+FV+RGB)~\cite{chen2017multi} & LIDAR+Mono&  96.52 & 89.56 & 88.94 & 96.02 & 89.05 & 88.38 & 71.29 & 62.68 & 56.56 \\
				 Frustum PointNets~\cite{qi2017frustum} & LIDAR & - & - & - & - & - & - & 83.76 & 70.92 & 63.65 \\
				\hline	\hline
				Ours (w/o clipping)-BirdNet & Mono & 55.70 & 38.42 & 37.20 & 51.27 & 37.41 & 30.28 & 8.43 & 4.26 & 3.12 \\
				Ours (w/o clipping)-MV3D & Mono & 61.4 & 46.18 & 44.20 & 59.23 & 45.46 & 41.72 & 46.42 & 38.70 & 25.35 \\
				Ours (clipping)-BirdNet & Mono & 89.4 & 68.3 & 64.3 & 80.34 & 51.20 & 46.7 & 12.24 & 10.70 & 8.64 \\
				Ours (clipping)-MV3D & Mono & 91.20 & 81.42 & 75.57 & 84.18 & 78.64 & 74.50 & 58.26 & 42.48 & 40.72 \\
				\hline
			\end{tabular}
			}
		\end{small}
		\caption{{\bf 3D detection performance:} Average Precision (AP$_{\text{3D}}$) (in \%) of 3D boxes on KITTI {\it validation} set. For Mono3D and 3DOP, we use the results with 3D box regression as reported in \cite{chen2017multi}}
		\label{tab:ap_3d_val}
		\vspace{-1em}
	\end{center}
\end{table*}

\begin{table}[t]
	\centering
	\resizebox{\columnwidth}{!}{
	\begin{tabular}{c c c c  c c c }
		\toprule
		  & \multicolumn{3}{c}{3D Object Detection} & \multicolumn{3}{c}{3D Object Localization} \\  	
	    \cmidrule(lr){2-4} \cmidrule(l){5-7} 
	 Method & Easy & Moderate & Hard & Easy & Moderate & Hard \\ 
		\midrule   
		BirdNet ~\cite{beltran2018birdnet}                  & 14.75           & 13.44         & 12.04          & 50.81            & 75.52           & 50.00       \\
MV3D ~\cite{chen2017multi}                     & 71.09           & 62.35          & 55.12          & 86.02            & 76.90           & 68.49           \\ 
AVOD ~\cite{ku2018joint}                    & 81.94           & 71.88          & 66.38          & 88.53            & 83.79           & 77.90           \\ \hline \hline
Ours (clipping)-BirdNet     & 10.01           & 9.42           & 7.20          & 46.01            & 65.31           & 41.45           \\ 
Ours (clipping)-MV3D     & 66.30           & 59.31           & 42.80          & 82.90            & 73.45           & 61.16           \\ 
Ours (clipping)-AVOD     & 77.24           & 61.52           & 52.30          & 84.40           & 78.41           & 72.20           \\ 
		\bottomrule
	\end{tabular}
	}
	\caption{Results on KITTI \textit{test} set for 3D Obj. Detection and Localization.}
\label{tab:kittitest}
\vspace{-1em}
\end{table}

\begin{table}[]
\resizebox{.95\columnwidth}{!}{
\scriptsize 
\begin{tabular}{c c c c c c }
\toprule
\multicolumn{1}{c}{Method} & \multicolumn{1}{c}{BB Opt.} & \multicolumn{1}{c}{Data} & \multicolumn{1}{c}{Easy} & \multicolumn{1}{c}{Mod.} & \multicolumn{1}{c}{Hard} \\ \midrule
Chabot et al.\ \cite{chabot2017deep}                      & 2D                                    & Mono                               & 96.40                              & 90.10                              & 80.79                              \\ 
Xu et al.\ \cite{xu2018multi}                            & 2D                                    & Mono                               & 90.43                              & 87.33                              & 76.78                              \\

AVOD \cite{ku2018joint}                                 & 3D                                    & LiDAR                              & 88.08                              & 89.73                              & 80.14                              \\

MV3D \cite{chen2017multi}                                  & 3D                                    & LiDAR                              & 90.37                              & 88.90                              & 79.81                              \\ \hline \hline
Ours(clipping)-MV3D                   & 3D                                    & Mono                               & 85.41                               & 81.72                               & 65.56                               \\ 
Ours(clipping)-AVOD                   & 3D                                    & Mono                               & 82.93                               & 83.51                               &  68.20                              \\ \bottomrule
\end{tabular}}
\caption{Performance on KITTI \textit{test} set for 2D Object Detection. BB Opt. indicates the optimization of bounding boxes in 2D or 3D.}
\label{tab:kitti2d}
\vspace{-1em}
\end{table}

\subsection{Quantitative results}

\noindent 
\textbf{BEV Object Detection:}
Table~\ref{tab:ap_loc_val} shows the results for 3D object localization of Bird's Eye View boxes on KITTI Benchmark. The 3D localization refers to the oriented object bounding boxes in the KITTI bird's eye view images. The evaluation is performed for both IoU = $0.5$ and IoU = $0.7$ where the performance is evaluated on Average Precision ($AP_{loc}$) for bird's eye view boxes. We observe that the proposed methods significantly outperform state-of-the-art results based on monocular (Mono3D) and stereo (3DOP) images. Specifically, the proposed method outperforms Mono3D by $27.9$\% on easy images and by $19.54$\% on hard examples with BirdNet, where the BEVs are obtained using BirdGAN trained on unclipped data. 

With the more restricted evaluation setting of IoU = $0.7$, both Mono3D and 3DOP suffer a significant drop in AP$_{\text{loc}}$ i.e.\ $\sim15$--$25$\%, while the drop for pipelines based on proposed network is more graceful at $\sim10$--$15$\% on an average. Similar trend is obtained for other compared methods as well. 

Further, when the networks are trained with clipped dataset, the improvement in AP$_{\text{loc}}$, in most cases, is $2$--$3$ times that of Mono3D and 3DOP. In fact, with clipped dataset, the performance of both BirdNet and MV3D are within $\sim5$--$10$\%, except on hard examples with IoU = $0.7$. The reason could be that while detecting examples with IoU = $0.7$ is a difficult task in itself \cite{chen2017multi}, the hard examples are heavily occluded and are relatively far in many cases. Since BirdGAN is trained specifically on close objects, the far away and heavily occluded objects are badly generated in the BEV image. 

Among the two base networks, BirdNet has nearly half the  AP$_{\text{loc}}$ of MV3D ($36.1$ \vs $60.13$). The drop in performances, compared to their respective baseline methods trained on real data, is more in case of BirdNet ($54.5$ to $36.1$) than MV3D ($76.67$ to $60.13$) showing that MV3D is better than BirdNet in the case of both real and generated data.

The proposed methods with clipped data perform $\sim10$-$25$\% better than corresponding networks trained with data without clipping. This shows that reducing noisy examples during training, allows GAN to learn a better mapping from 2D image domain to BEV image. While this leads to discarding data at training time, overall the less noisy training improves performance at test time by learning better quality BEV generators.

\noindent
\textbf{3D Object Detection:} The results for 3D Object Detection \ie detection of 3D bounding boxes are shown in Table \ref{tab:ap_3d_val}. The performance is evaluated by computing IoU overlap with ground-truth 3D bounding boxes using Average Precision ($AP_{3D}$) metric. With IoU = $0.25$, the Mono3D and 3DOP outperform the BirdNet and MV3D trained on unclipped dataset. However, with clipped dataset, MV3D performs better than both Mono3D and 3DOP which are based on monocular images. With IoU $0.5$ and $0.7$, the proposed pipeline with or without clipped data still outperforms Mono3D and 3DOP. In fact, MV3D on generated data performs $\sim30$\% (w/o clipping) to $\sim40$\% (clipped) better on IoU $0.25$ and $0.5$ than Mono3D and 3DOP. Except MV3D and Frustum PointNets, the proposed MV3D with clipped data still outperforms existing methods which explicitly use real 3D data like  VeloFCN \cite{li2016vehicle}.

\begin{figure*}
\begin{center}
\resizebox{\linewidth}{!}{
\begin{tabular}{ccc|cc}
\includegraphics[width=0.2\linewidth,trim = 0mm 0mm 0mm 65mm, clip]{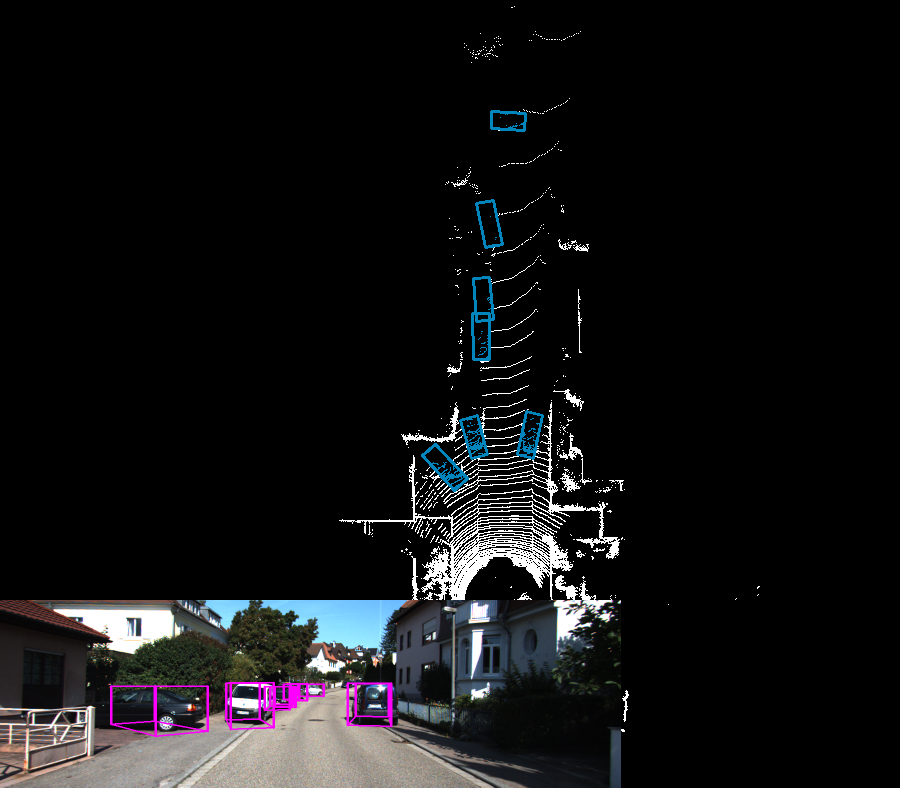}
&
\includegraphics[width=0.2\linewidth,trim = 0mm 0mm 0mm 65mm, clip]{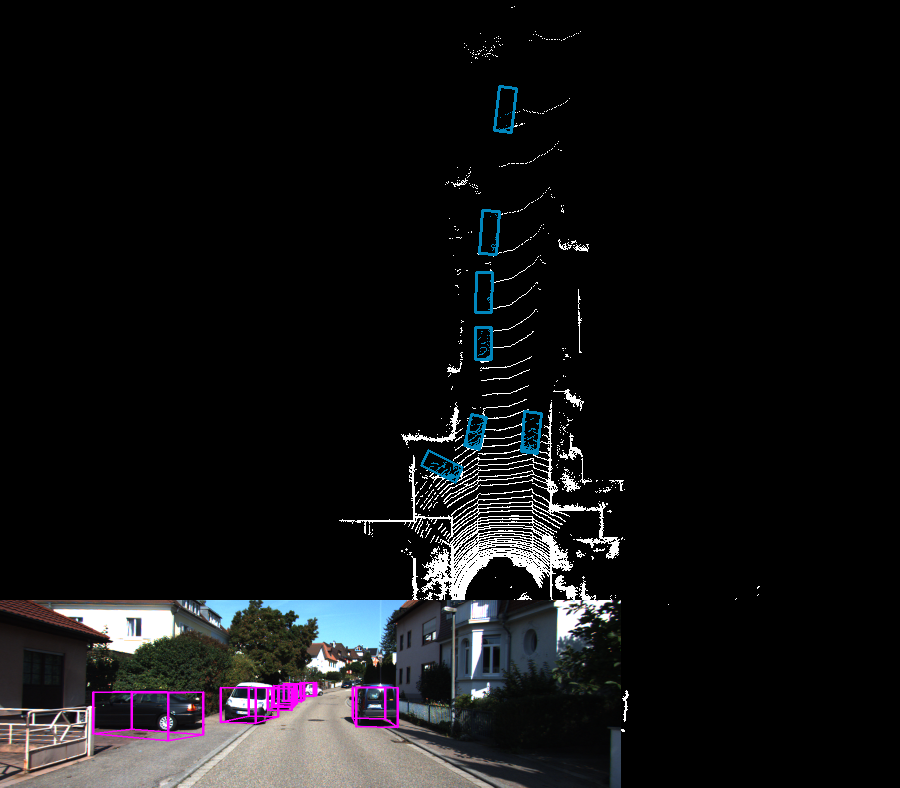}
&
\includegraphics[width=0.2\linewidth,trim = 0mm 0mm 0mm 65mm, clip]{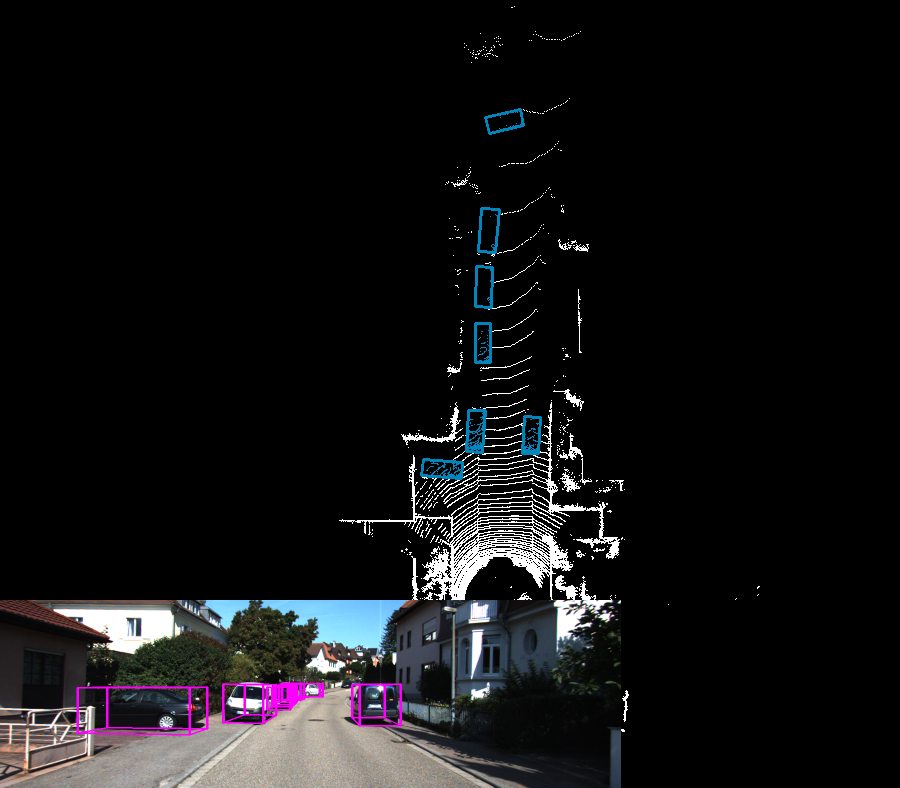}
&
\includegraphics[width=0.2\linewidth,trim = 0mm 0mm 0mm 33mm, clip]{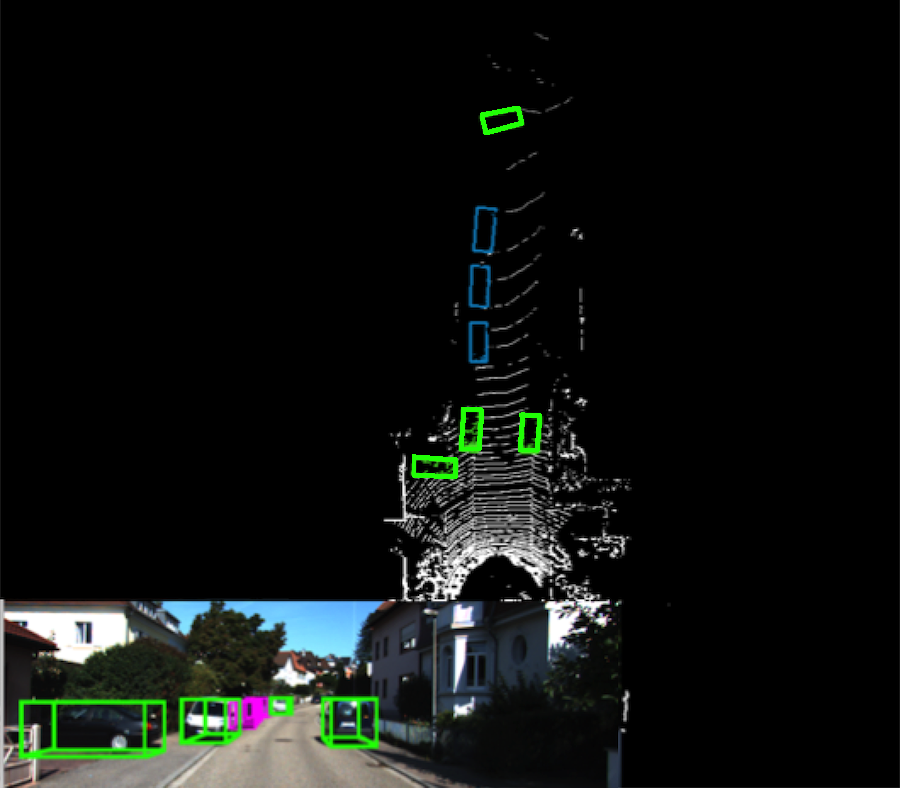}
&
\includegraphics[width=0.2\linewidth,trim = 0mm 0mm 0mm 38mm, clip]{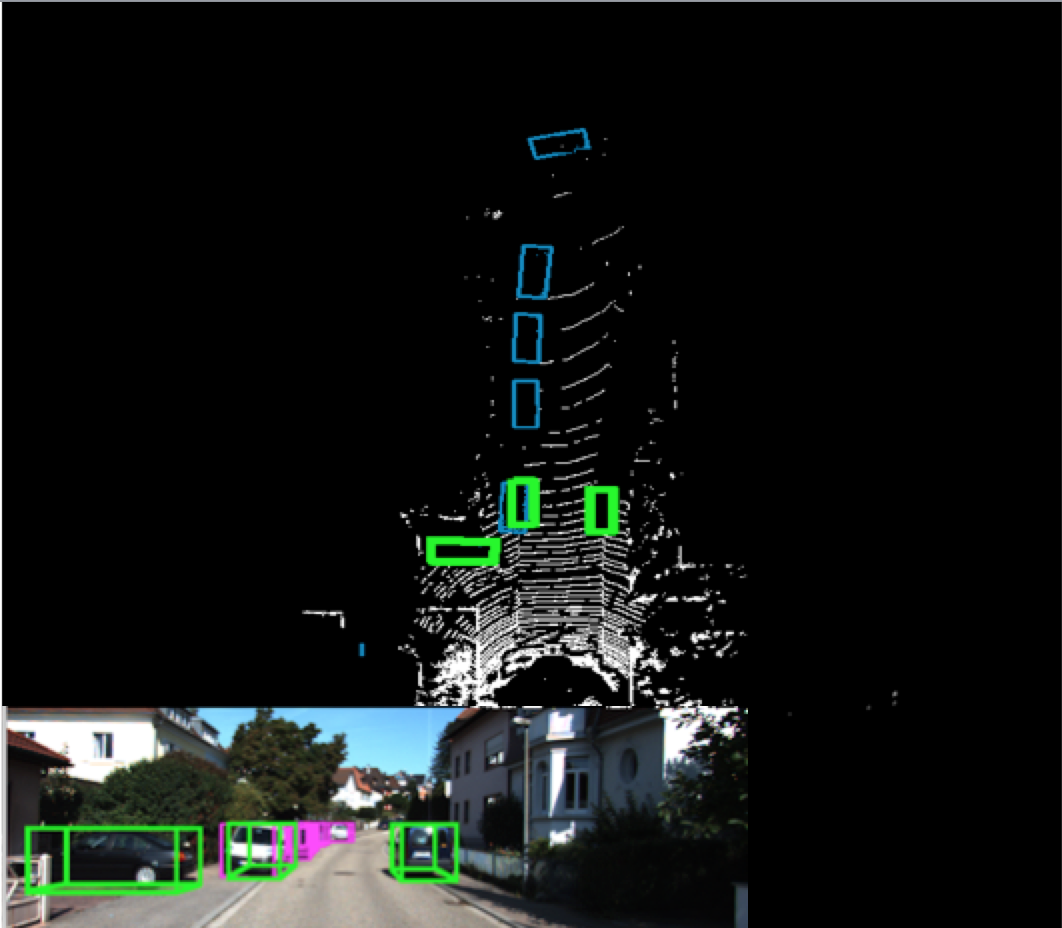}
\\
\includegraphics[width=0.2\linewidth,trim = 0mm 0mm 0mm 70mm, clip]{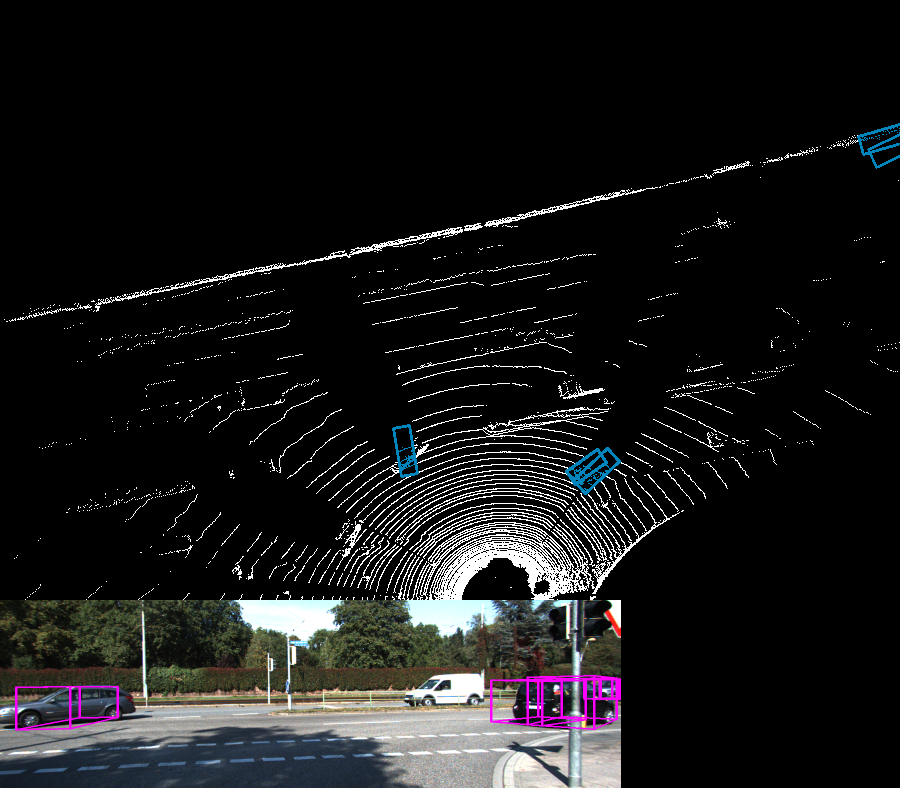}
&
\includegraphics[width=0.2\linewidth,trim = 0mm 0mm 0mm 70mm, clip]{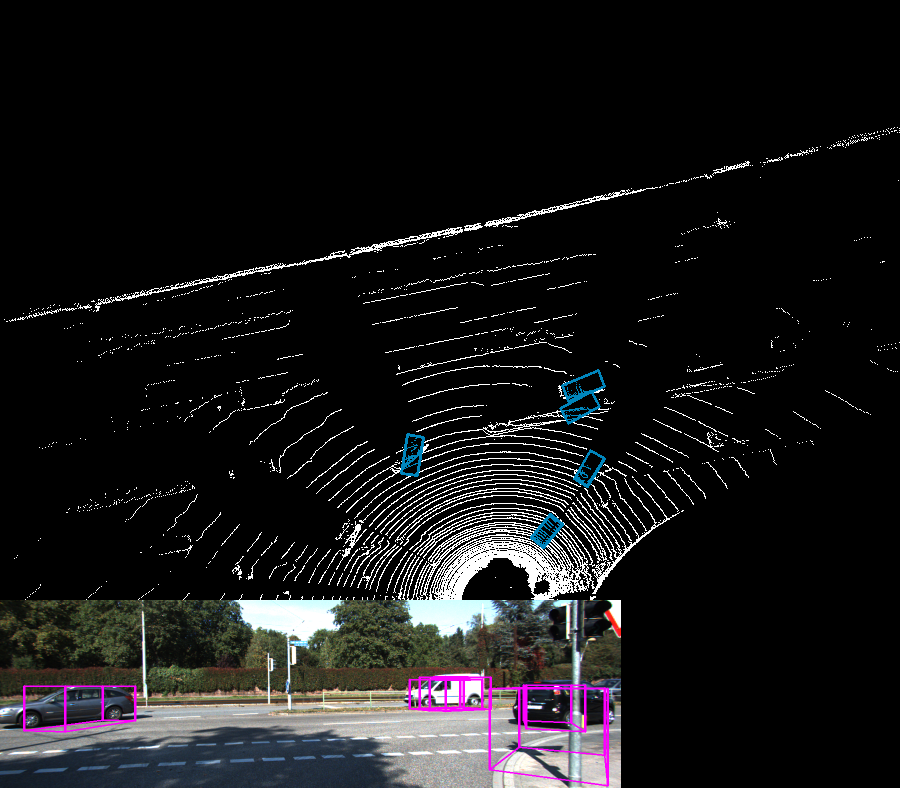}
&
\includegraphics[width=0.2\linewidth,trim = 0mm 0mm 0mm 70mm, clip]{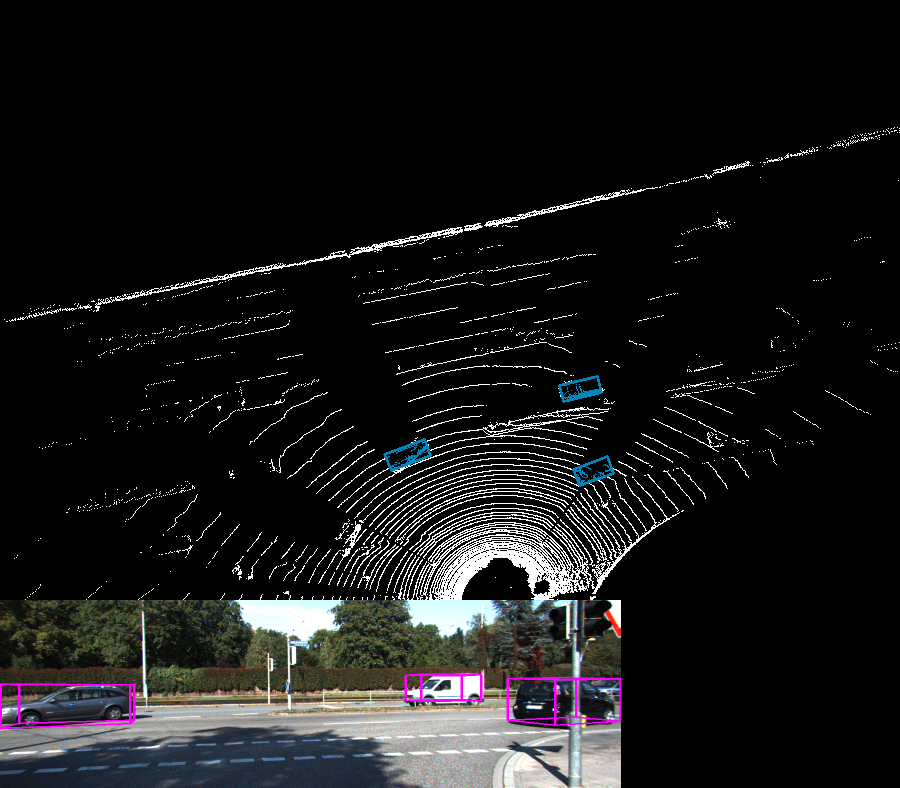}
&
\includegraphics[width=0.2\linewidth,trim = 0mm 0mm 0mm 33mm, clip]{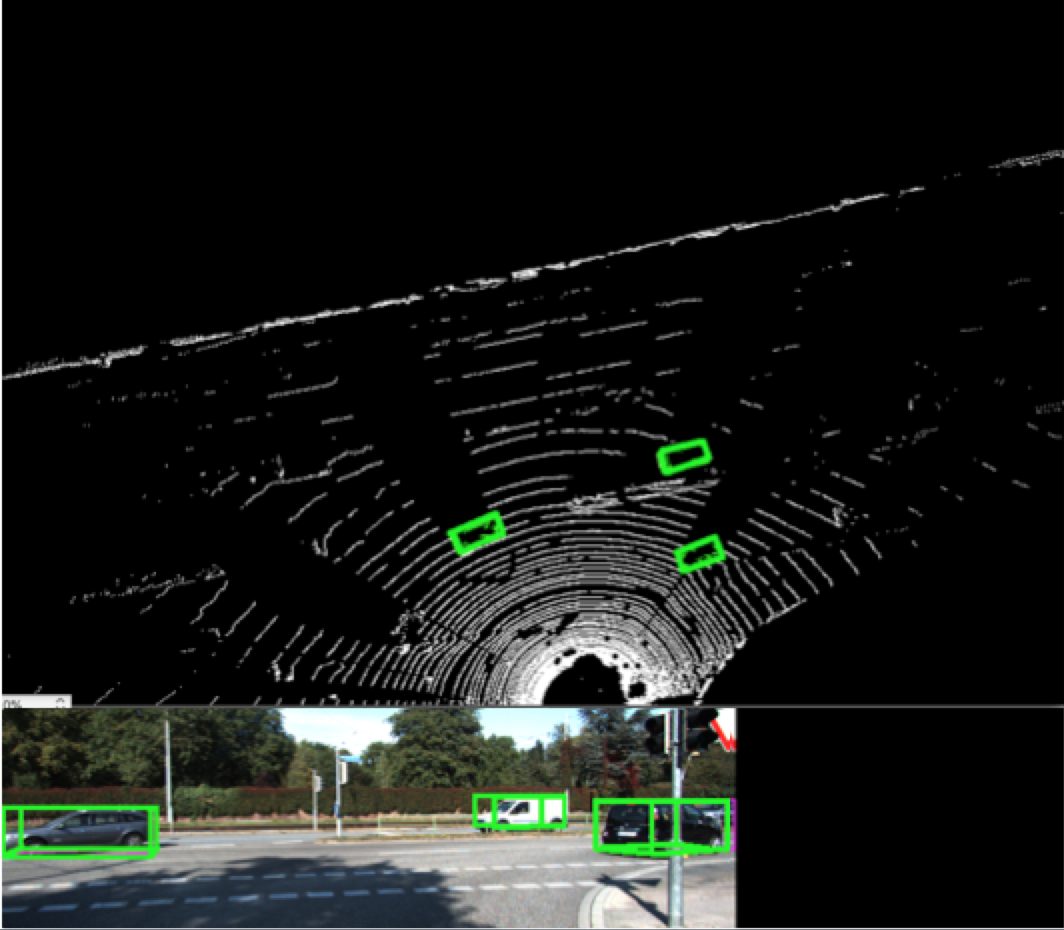}
&
\includegraphics[width=0.2\linewidth,trim = 0mm 0mm 0mm 77mm, clip]{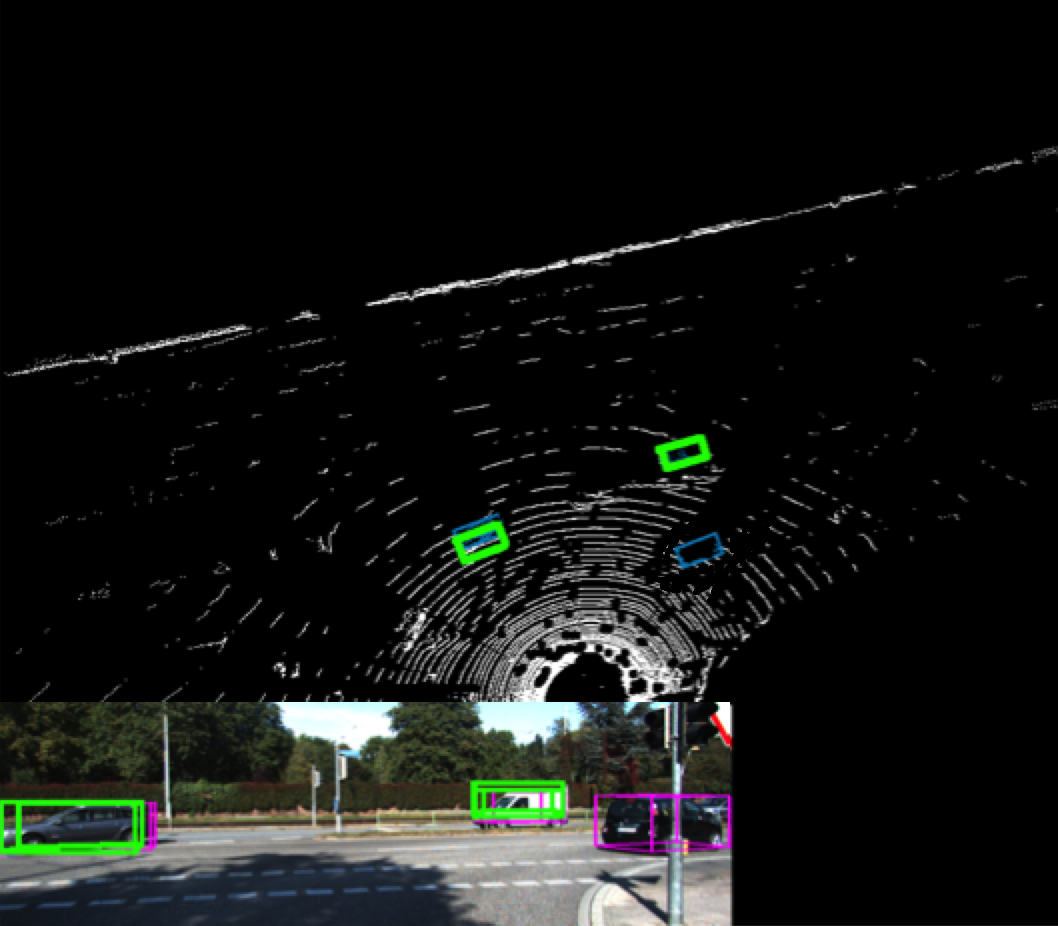} \\
3DOP~\cite{chen20153d} & VeloFCN~\cite{li2016vehicle} & MV3D\cite{chen2017multi} & Ours(clipped)-MV3D & Ours(clipped)-BirdNet\\
\end{tabular}
}

\caption{\textbf{Qualitative Results} on 3D Detection for various techniques against the proposed pipelines. The results for Ours(clipped)-BirdNet and Ours(clipped)-MV3D show the computed results in green. The Bird's Eye View images shown in the results for the proposed pipelines are generated using BirdGAN. To highlight the quality of detections using the proposed techniques, the detections using default MV3D are also marked in the RGB (pink) and BEV (blue) images for Ours(clipped)-MV3D and Ours(clipped)-BirdNet. }
\label{fig:vis}
\end{center}
\vspace{-1em}
\end{figure*}

Similar to the case of 3D localization, we observe that the networks trained with clipped data achieve a significant increase in the $AP_{3D}$ as compared to the networks trained on unclipped data. The performance increase is $\sim30$\% on IoU $0.25$ for both BirdNet and MV3D. The results indicate that with clipped data, the detected bounding boxes are close to the actual object. However, with IoU of $0.7$, the increase in performance for networks trained on unclipped data \vs clipped data is reduced to $\sim4$\% for BirdNet and $\sim8$\% for MV3D. This indicates that BEV generated with clipped data allows learning of models that have a larger number of bounding boxes closer to the ground-truth annotations, which also have a larger overlap with them. The increase for MV3D is interesting as it uses LIDAR Front View which is also generated via depth map obtained only from the RGB image. 

We also observe an increase of $\sim15$\% in case of \emph{hard} examples with an IoU of $0.7$ for MV3D with clipped \cf unclipped data. This again supports our hypothesis that clipping the data and training on closer objects, that have better visibility in the RGB image and lesser occlusions, helps the BirdGAN to learn a better mapping from RGB image to the Bird's Eye View image. 

\noindent
\textbf{Generalization Ability:} In Table \ref{tab:kittitest}, the results on KITTI test set are shown. We can observe that the performance with proposed method is again close to the performance of the base network. Additionally, to demonstrate that the proposed method can be used as a drop-in replacement, we show the results with another base architecture, AVOD \cite{ku2018joint} which uses the same BEV as MV3D. It also generates 3D anchor grids, for which we provide the point clouds generated using the method discussed in Section \ref{subsec:mv3darch}. We observe that the performance with the proposed method is very close to AVOD's performance with real 3D data, and that it outperforms MV3D, with real 3D data, on three out of six cases, i.e.\ on localization, moderate and hard, and detection easy. Thus the proposed method is general and can work with different state of the art 3D networks giving competitive performance.
\\
\noindent
\textbf{2D Object Detection:} In Table \ref{tab:kitti2d}, the results on KITTI test set for 2D detections are shown. It can be observed that even with entirely generated data the method also performs close to the base networks for 2D object detection. 

\subsection{Qualitative Results}
Figure \ref{fig:vis} shows some qualitative results for object detection in 3D, on actual BEV images for compared methods (first three columns) and on generated BEV images for the proposed methods. For comparison we chose 3DOP which is based on stereo images, VeloFCN is based on LIDAR and MV3D, which is based on multi modal fusion of LIDAR, BEV and RGB image. In the first row, we can observe that Ours(clipped)-MV3D detects four cars in the scene while missing the occluded car. However, Ours(clipped)-BirdNet detects three cars closer to the camera while misses out on the car at the back, which is visible but is far away from the camera. We can also observe that the BEV image and the marked detections in Ours(clipped)-MV3D, which uses generated BEVs, are very close to the full MV3D with actual BEV images. However, for Ours(clipped)-BirdNet, the detections in BEV do not overlap completely with the object explaining the drop in performance when higher IoUs are considered with BirdNet for 3D Object localization and detection (Table \ref{tab:ap_loc_val} and \ref{tab:ap_3d_val}). 

The second row shows a simpler case, where only three cars are present in the image. It can be seen that Ours(clipping)-MV3D is able to detect all three of the cars with high overlap with default MV3D. However, with Ours(clipping)-BirdNet, only two cars are detected. The third car (right most) is not detected potentially because BirdNet is highly sensitive to occlusion and the presence of the pole in front of that car is confusing it. 

When compared to the generated BEV for MV3D ($M+2$ channel) and the ground-truth (row 2, first three figures), the generated BEV for BirdNet (only $3$ channels) has missing regions. These regions have better reconstruction in the first case, potentially because the BEV consists of multiple height channels, which might be allowing it to distinguish the region where car ends while the pole continues (from the perspective of converting RGB to BEV image). This case also shows the effectiveness of the BirdGAN in generating close to actual Bird's Eye View images where it respected the occlusions due to cars and large objects, which appear as empty blank region in the BEV images, e.g.\ behind the cars in the generated as well as actual BEV image.  

\begin{table*}
	\centering
	\resizebox{\linewidth}{!}{
	\begin{tabular}{c c c  c c c  c c c  c c c  c c c}
		\toprule
		I & D & H & \multicolumn{3}{c}{BirdNet\cite{beltran2018birdnet}} & \multicolumn{3}{c}{Ours(clipped)-BirdNet} & \multicolumn{3}{c}{MV3D\cite{chen2017multi}} &
		\multicolumn{3}{c}{Ours(clipped)-MV3D}\\  	
		\cmidrule(lr){4-6} \cmidrule(l){7-9} \cmidrule(l){10-12} \cmidrule(l){13-15} 
		&   &  & Easy & Moderate & Hard & Easy & Moderate & Hard & Easy & Moderate & Hard & Easy & Moderate & Hard \\ 
		\midrule   
		\checkmark &  \checkmark  &  \checkmark  & 72.32 & 54.09 & 54.50 & 68.2 & 42.1 & 36.1 & 86.18 & 77.32 & 76.33 & 81.32 & 68.40 & 60.13 \\
		\midrule  
		\checkmark & &  & 55.04 & 41.16 & 38.56 & 51.32 & 28.21 & 18.65 & 68.20 & 65.66 & 62.14 & 62.10 & 45.36 & 42.64 \\
		& \checkmark &  & 70.94  & 53.00 & 53.30  & 65.50  & 38.42 & 33.20 & 75.30 & 69.45  & 68.32 & 72.20 & 58.49  & 48.24 \\
		& & \checkmark & 69.80 & 52.90 & 53.69 & 64.70 & 36.10 & 32.44 & 75.11 & 73.10 & 67.50 & 72.45 & 59.22 & 48.68 \\         
		\bottomrule
	\end{tabular}
	}
	\caption{Performance on Bird Eye's View Detection (AP$_{\text{loc}}$) on the KITTI \textit{validation} Set for Car using different channel of Bird Eye's View image as an input. I represents Intensity, D represents Density and H represents Height channels of a Bird's Eye View.}
	\label{tab:ablation}
	\vspace{-2em}
\end{table*}

\begin{table}
	\centering
	\resizebox{\columnwidth}{!}{
	\begin{tabular}{c c c c  c c c }
		\toprule
		  & \multicolumn{3}{c}{3D Object Localization (AP$_{\text{loc}}$)} & \multicolumn{3}{c}{3D Object Detection (AP$_{\text{3D}}$)} \\  	
	    \cmidrule(lr){2-4} \cmidrule(l){5-7} 
	 Method & Easy & Moderate & Hard & Easy & Moderate & Hard \\ 
		\midrule   
		 BirdNet$_{fus}$ & 75.64 & 57.09 & 55.48 & 21.20 & 17.21 & 12.10 \\
		 MV3D$_{fus}$ & 89.24 & 79.12 & 73.10 & 75.31 & 64.70 & 52.80 \\
		\bottomrule
	\end{tabular}
	}
	\caption{Performance of Deep Fusion of outputs from networks trained on real and generated BEV images for 3D Object Localization and Detection on KITTI. }
	\vspace{-1em}
	\label{tab:ablationfusion}
\end{table}

\subsection{Ablation Studies}
We first study the impact of various channels within BEV on 3D object detection and localization. Table \ref{tab:ablation} shows the results on using each of the channels of BEV image separately. We observe that by removing the density and height channels, there is a high drop in performance for both BirdNet and MV3D. In case of BirdNet, using only the density or the height channel provides performance closer to the case when all three channels have been used. In contrast, for MV3D, the drop in relative performance is comparatively higher ($\sim5$-$12$\%) when using density or height channel alone as compared to BirdNet where the relative drop is $\sim2$-$5$\%. This indicates that both the channels encode distinctive and/or complementary information which results in significant boost when all the channels are combined.

Next, we experiment with using generated data along with the real data to analyze if the generated data can improve the detection and localization performance by augmenting the real data. We first try merging the ground-truth training images and the generated BEV image to make a common dataset and train BirdNet and MV3D with it. However, we observe that the performance drops about $12$\% on an average in this setting on the easy, moderate and hard examples. We hypothesize that this could be because the network was not able to optimize over the combined dataset which includes two BEVs for the same image, one real and one generated. The two BEVs would have different statistics and might thus confuse the detector when used together for training.

However, since the networks independently perform well, we perform their deep fusion, similar to that in MV3D. We combine their outputs with a join operation \eg concatenation or mean, and feed that to another layer before prediction, \ie we add another layer whose input is the join of the outputs from the two networks given by
\begin{align}
f_{BirdNet_{fus}} = & f_{BirdNet} \oplus f_{(BirdGAN+BirdNet)} \\
f_{MV3D_{fus}} = & f_{MV3D} \oplus f_{(BirdGAN+MV3D)} 
\end{align}
where $f_{BirdNet}$ and $f_{MV3D}$ are networks pretrained on ground-truth data while $f_{(BirdGAN+BirdNet)}$ and $f_{(BirdGAN+MV3D)}$ are BirdNet and MV3D networks pretrained on generated BEV. $f_{BirdNet_{fus}}$ and $f_{MV3D_{fus}}$ is the combination of the respective outputs using the join operation $\oplus$, which is a mean operation in our case. The results are shown in Table \ref{tab:ablationfusion}. We observe that deep fusion improves the performance of the base network for both 3D localization (Table \ref{tab:ap_loc_val}) and 3D detection (Table \ref{tab:ap_3d_val}) by $2$-$5$\% for easy and moderate examples. Interestingly, the addition of generated data for 3D object localization on easy examples, provides an AP$_{\text{loc}}$ of $89.24$, which is higher than that of a state-of-the-art method, Frustum PointNets ($88.16$). However, in case of hard examples, the performance drops. This could be due to the fact that the hard examples contain heavily occluded objects and hence the introduction of generated BEVs reduces the performance \cf the network trained on ground-truth annotations. While in this, one of the networks uses LIDAR data, the results are interesting, as we can generate data from the available training data, and use that to improve the performance without needing additional training samples. We believe this is one of the first time performance improvment has been reported by augmenting with generated data on a real world computer vision problem.

\section{Conclusion}
We demonstrated that using GANs to generate 3D data from 2D images can lead to  performances close to the state-of-the-art 3D object detectors which use actual 3D data at test time. The proposed method outperformed the state-of-the-art monocular image based 3D object detection methods by a significant margin. We proposed two generation mechanisms to work with two different recent 3D object detection architectures, and proposed training strategies which lead to high detection performances. We also gave ablation studies and compared the results from the architectures using real 3D data \vs generated 3D data, at train and test time. We also showed that late fusion of networks trained with real and generated 3D data, respectively, improves performance over both individually. We believe it is one of the first time that results have been reported where training data augmentation by generating images leads to improved performance for the challenging task of 3D object detection. The setting used in our experiments is very practical as urban environments remain relatively similar in local areas. Hence the RGB image to 3D BEV image generator once trained, say in parts of a certain city, can be used with good generalization in different areas of the same and or nearby cities. An avenue for future work is to move towards more challenging setting with unknown camera parameters (\eg the lines of \cite{ZhuangIROS2019}).
\\ \ \\
\noindent
\textbf{Acknowledgements.} This work was partly funded by the MOBI-DEEP ANR-17-CE33-0011 program.
{
 \small
\bibliographystyle{IEEEtran}
\bibliography{IEEEabrv,egbib}
}

\end{document}